%% file: main_new.tex
\documentclass[conference]{IEEEtran}

\usepackage{blindtext}
\usepackage{eso-pic}
\IEEEoverridecommandlockouts
\usepackage{cite}
\usepackage{amsmath,amssymb,amsfonts}
\usepackage{algorithmic}
\usepackage{graphicx}
\usepackage{textcomp}
\usepackage{xcolor}

\usepackage[export]{adjustbox}
\usepackage[utf8]{inputenc} 
\usepackage[T1]{fontenc}    
\usepackage{hyperref}       
\usepackage{url}            
\usepackage{booktabs}       
\usepackage{amsfonts}       
\usepackage{nicefrac}       
\usepackage{microtype}      
\usepackage{xcolor}         
\usepackage{amsmath,amssymb,multicol,latexsym}
\usepackage{pgfplots,pgfplotstable}
\pgfplotsset{compat=1.16}
\usepackage{tikz}
\usepackage{subcaption}
\usepackage{cite}
\usetikzlibrary{shapes,
 	automata,
 	arrows,
 	chains,
 	matrix,
 	backgrounds,
 	fit,
 	patterns,
 	decorations.markings, 
 	svg.path, 
 	shapes.multipart, 
 	shapes.symbols, 
 	external, 
 	shadows, 
 	positioning, 
 	calc, 
 	decorations, 
 	scopes,
 	patterns,
 	patterns.meta,
 	bending}
 \usetikzlibrary{external}

\usepackage{hyperref}
\usepackage{cleveref}
\usepackage{flushend}

\DeclareMathAlphabet{\mathcal}{OMS}{cmsy}{m}{n}

\usepackage{url}
\usepackage{graphicx}
\usepackage{svg}
\usepackage{tkz-kiviat,numprint,fullpage}
\usepackage{booktabs}
\usepackage{multirow}

\usepackage{adjustbox}

\Crefname{equation}{Eq.}{Eqs.}
\Crefname{figure}{Fig.}{Figs.}
\Crefname{tabular}{Tab.}{Tabs.}
\usepackage{tabu}
\usepackage{anysize} 
\graphicspath{{figures/}}
\def\BibTeX{{\rm B\kern-.05em{\sc i\kern-.025em b}\kern-.08em
    T\kern-.1667em\lower.7ex\hbox{E}\kern-.125emX}}
    
\usepackage{eso-pic}

\begin{document}
\bstctlcite{IEEEexample:BSTcontrol}
\title{\vspace*{1cm} Fingerprint of a Traffic Scene:\\an Approach for a Generic and Independent Scene Assessment\\
}

\author{
\IEEEauthorblockN{Maximilian~Zipfl$^{\dagger}$}
\IEEEauthorblockA{\textit{FZI Research Center for Information Technology} \\
Karlsruhe, Germany \\
zipfl@fzi.de}
\and
\IEEEauthorblockN{Barbara~Schütt$^{\dagger}$}
\IEEEauthorblockA{\textit{FZI Research Center for Information Technology} \\
Karlsruhe, Germany \\
schuett@fzi.de}
\and
\IEEEauthorblockN{J.~Marius~Zöllner}
\IEEEauthorblockA{\textit{FZI Research Center for Information Technology} \\
\textit{\&}\\
\textit{Karlsruhe Institute of Technology}\\
Karlsruhe, Germany \\
zoellner@fzi.de}
\and
\IEEEauthorblockN{Eric~Sax}
\IEEEauthorblockA{\textit{FZI Research Center for Information Technology} \\
\textit{\&}\\
\textit{Karlsruhe Institute of Technology}\\
Karlsruhe, Germany \\
sax@fzi.de}
\thanks{$^{\dagger}$  Both authors contributed equally to this work as first authors.}
}

\maketitle
\begin{abstract}
A major challenge in the safety assessment of automated vehicles is to ensure that risk for all traffic participants is as low as possible.
A concept that is becoming increasingly popular for testing in automated driving is scenario-based testing.
It is founded on the assumption that most time on the road can be seen as uncritical and in mainly critical situations contribute to the safety case.
Metrics describing the criticality are necessary to automatically identify the critical situations and scenarios from measurement data.
However, established metrics lack universality or a concept for metric combination.
In this work, we present a multidimensional evaluation model that, based on conventional metrics, can evaluate scenes independently of the scene type. Furthermore, we present two new, further enhanced evaluation approaches, which can additionally serve as universal metrics. The metrics we introduce are then evaluated and discussed using real data from a motion dataset.
\end{abstract}


\input{content/01_introduction.tex}


\input{content/02_related_work}

\input{content/03_implementation}

\input{content/05_evaluation}

\input{content/06_conclusion}

 \section{Acknowledgement}
 The research leading to these results is funded by the German Federal Ministry for Economic Affairs and Climate Action within the projects “Verifikations- und Validierungsmethoden automatisierter Fahrzeuge im urbanen Umfeld" and “SET Level – Simulation-based Development and Testing of Automated Driving" both projects from the PEGASUS family, based on a decision by the Parliament of the Federal Republic of Germany. The authors would like to thank the consortium for the successful cooperation.

 This work was also partially supported by the Intel Collaborative Research Institute for Safe Automated Vehicles (ICRI-SAVe).

\bibliographystyle{IEEEtran}
\bibliography{references, bst_info}

\end{document}

%% file: content/01_introduction.tex
\section{Introduction}
Highly automated driving has emerged as one of the automotive industry's most significant strategic research areas. It promises to increase driving safety and reduce traffic accidents, offering drivers greater convenience and comfort by relieving them of stressful tasks such as traffic jams \cite{fagnant_preparing_2015}.

The evaluation of traffic scenes and traffic scenarios plays an essential role in verifying and validating highly automated driving functions. Due to the high effort required to cover a driving function statistically \cite{maurer_release_2016, pretschner_tests_2021}, more and more research efforts are being made in favor of scenario-based testing.
There are two different approaches for generating scenes and scenarios. The knowledge-based approach derives scenarios from guidelines, ontologies, and rules. The data-based approach refers to recordings and accident data from which scenarios or scenes are extracted.
In general, traffic scenes are recorded with static recording devices \cite{zipfl_traffic_2020,bock_ind_2019,zhan_interaction_2019} or with sensorized vehicles driving on the road \cite{chang_argoverse_2019,geiger_vision_2013,caesar_nuscenes_2020} and then are analyzed for important scenarios depending on the particular use case. These scenarios can then be used to test driving functions, for instance, in a simulation.
The decisive role here, however, is how the scenario is evaluated. In most cases, criticality metrics are used. These describe whether a scene or a scenario contains dangerous elements.
One problem is that many criticality metrics focus on a specific feature. Also, not all criticality metrics are applicable to all scenes or require specific conditions.
Furthermore, often, only whole scenes or only single entities in a scene are considered separately.
\begin{figure}[t!]
    \centering
       \includegraphics[width=0.92\linewidth]{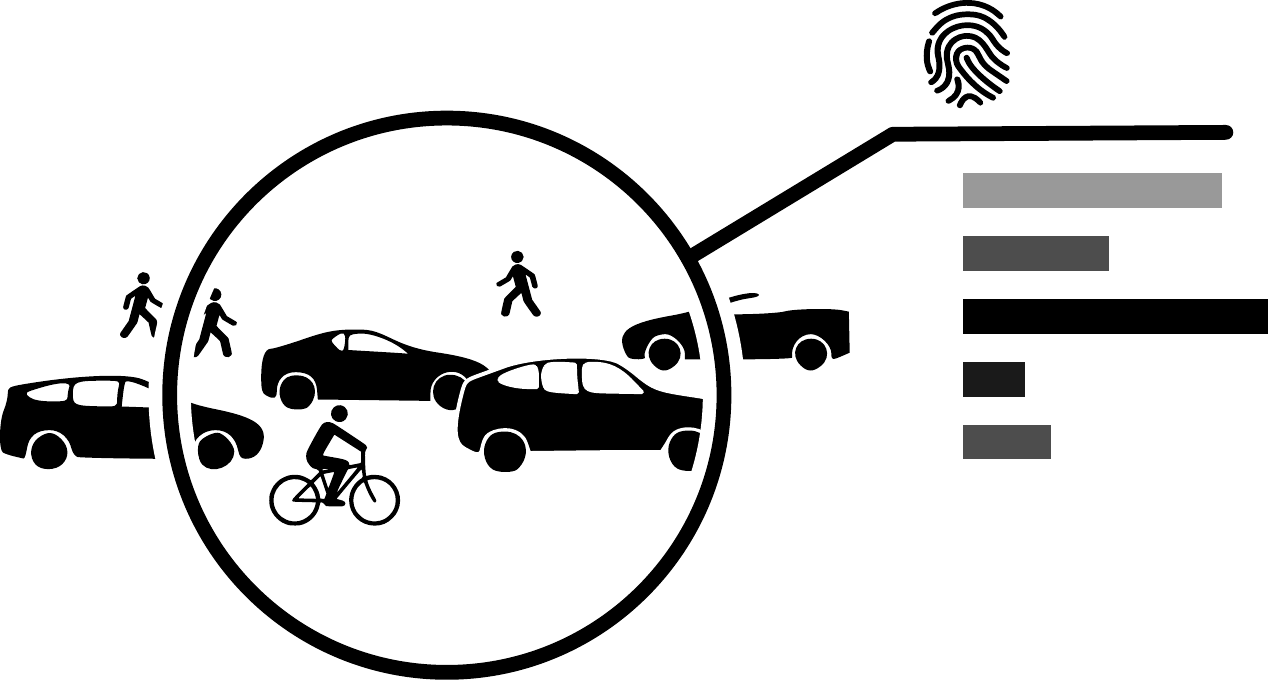}
\caption{Fingerprint: A traffic scene is described by metrics from different domains.}
\label{fig:approach_top_right}
\vspace{-1ex}
\end{figure}
This paper presents a framework for comparing traffic scenes based on motion datasets using various metrics. In addition, we evaluate several established metrics concerning different traffic scene types and suggest improvements. 
By combining and extending universal metrics to a scene representation, the applicability in reality and associated challenges should be possible. 
One of the goals of this work is to overcome the problem of the limited applicability of scenario types, as is the case with many metrics in use, e.g., only measurable at intersections \cite{westhofen_criticality_2022}.

This paper's novelty and main contribution is a new way of criticality metrics evaluation. 
Thus, we
\begin{itemize}
    \item propose an adjusted version of the traffic quality and safety potential metric for universal usage in all scenario types and
    \item a new way of metric visualization and evaluation using Kiviat diagrams (spider charts) and the area of criticality.
\end{itemize}

This paper is structured as follows: In \Cref{sec:sota}, we provide a review of existing criticality metrics and related work. In \Cref{sec:approach} and \Cref{sec:implementation}, we give a detailed definition of our improved metrics, the framework, and how it is implemented concretely. In \Cref{sec:evaluation}, properties of the combination approach are evaluated with examples. Finally, in \Cref{sec:conclusion}, we conclude this contribution.

%% file: content/02_related_work.tex
\section{Related Work}
\label{sec:sota}
In the context of this work, the terms \textit{scenario} and \textit{scene} are used as defined by the standard ISO/PAS 21448 \cite{technical_committee_isotc_22_road_vehicles_subcommittee_sc_32_electrical_and_electronic_components_and_general_system_aspects_isopas_nodate}. 
A \textit{Scene} is a snapshot of all dynamic objects, actors, and their observable states, as well as the scenery, e.g., weather or lane network.
A \textit{Scenario} is a chronological sequence of individual scenes and their transitions.

\subsection{Metric Classification}
Metrics are needed to determine the criticality of a scene or scenario \cite{schutt_application_2022}, \cite{baumann_automatic_2021}.
However, metrics can be used in different contexts throughout the development and testing of automated vehicles.

\subsubsection{Resolution based classification}
Schütt \textit{et al.} \cite{schutt_taxonomy_2022} suggest an extendable quality metrics taxonomy consisting of different domains of interest within the automotive testing domain, i.e., (simulation) model quality, a system under test quality, and scenario quality.
The quality in these domains, which also includes criticality, can be examined at different levels of resolution.
The highest resolution is nanoscopic and can measure quality for atomic models or for each time step.
The next level of resolution is microscopic quality and consists of quality observed for coupling mechanisms, e.g., middleware, scenarios, or time intervals.
The highest level of resolution is macroscopic quality and  measures quality, e.g., parameter coverage over sets of scenarios or the quality of coupled systems.
\subsubsection{Temporal-spatial based classification}
Mahmud \textit{et al.} \cite{mahmud_application_2017} propose different categories of proximal surrogate indicators: temporal-based metrics, distance-based metrics, deceleration-based metrics, and others.
Most common criticality metrics are based on temporal or spatial proximity, as they assume that the closer two traffic participants are to each other, the more likely they are to collide.
Metrics of this type usually indicate a conflict if the resulting value lies beneath a predetermined threshold.

\subsection{Criticality Metrics}
The majority of metrics are binary, meaning they can only be measured between two traffic participants and only indicate how critical a situation is between those two.
In the worst case, critical situations can stay undetected because not every actor within a scene was considered.
These binary criticality metrics are cheap to calculate and only need partial information about a scene.

\subsubsection{Time-based and Distance-based Metrics}
The most straightforward metric is the Euclidean distance between two vehicles.
It is easy to calculate since it only needs positional data. Nevertheless, it fails to give a reliable criticality measure in some scenarios, e.g., two participants driving next to each other in the same direction.
Another of the most popular criticality metrics is time to collision (TTC) \cite{junietz_metrik_2017}.

\begin{equation}
    TTC = \frac{d}{\nu_{f} - \nu_{l}}
\end{equation}
where $d$ is the distance between a leading and the following vehicle and $\nu_{f}$ and $\nu_{l}$ are the respective velocities.
A downside of TTC is that it can only be used for car-following scenarios.
Other situations, e.g., intersection scenarios, lead to unusable results.
Several approaches for TTC-related metrics were proposed to solve this problem of limited domains.
One of them is the worst TTC \cite{wachenfeld_worst-time--collision_2016}, which evaluates the criticality between two actors should they choose the most adverse intersecting trajectories possible. 

On the other hand, metrics like post-encroachment-time (PET), encroachment-time (ET), gap time (GT), or trajectory distance (TJ) are only suitable for intersection scenarios \cite{allen_analysis_1978}.
PET computes the time between two actors passing a point or area of intersection. ET announces how long the first of both actors spend time on the point or area of intersection and, hence, is encroaching upon the other vehicle's trajectory.
GT is highly related to PET. PET is usually computed after the simulation was executed or traffic was observed, whereas GT is calculated online and can be seen as a prediction of PET.
TJ is the distance along the trajectories of both actors until the intersection is reached.
According to Schütt \textit{et al.} \cite{schutt_taxonomy_2022}, all mentioned metrics except PET are nanoscopic metrics, since they can be calculated for each time step. 
PET is the only microscopic metric that needs a time series for calculation.

\subsubsection{Highway Traffic Quality}
Hallerbach \textit{et al.} \cite{hallerbach_simulation-based_2018} propose a method to evaluate the criticality of the traffic on a highway section. The focus is not solely on the ego vehicle and one other traffic participant but on different domains of interest, including larger road sections, traffic around the ego vehicle, and the ego vehicle itself.
The calculation of traffic quality is divided into four parts with differently weighted coefficients, which sum up to a total criticality score.
The first part, the macroscopic metric, calculates the traffic density with the help of the traffic flow rate and the average travel velocity.
Additionally, the microscopic metric considers the velocity deviation and the average velocity on a highway section around the ego vehicle.
The third part of the traffic quality term, the nanoscopic metric, is centered on close-range interactions inside a circle around the ego vehicle with a smaller radius, and its calculation is based on velocity deviation and mean value within it.
The last part is the individual metric and concentrates on the ego vehicle's mean velocity and standard deviation of the acceleration.
Finally, a supervised training algorithm finds optimal weights as coefficients for all four terms and has to be repeated for different scenes.

\subsubsection{Safety Force Field}

The idea behind the safety force field (SFF) model \cite{nister_introduction_2019, nister_safety_2019} is to avoid collisions by not contributing to an unsafe environment since an ego vehicle cannot guarantee a safe journey when other actors share the roads.

Similar to the RSS model \cite{shalev-shwartz_formal_2018}, the SFF is a model for evaluating the driving situation in which the highly automated vehicle currently operates to make a safe decision based on this model.  
The states of all traffic participants are considered and how they will behave in the near future to assess the situation. For this purpose, the area over time that a vehicle will occupy is calculated, referred to as the \emph{claimed set} (compare \Cref{fig:claimed-set}). An optimal safety procedure can be calculated based on the geometric overlaps of the occupied sets of different traffic participants.

%% file: content/03_implementation.tex
\section{Approach and Framework}
\label{sec:approach}
For metrics to be evaluated, they must be applicable to traffic scenes (see \Cref{fig:approach_top_right}). For this purpose, we present a calculation framework that loads the necessary data from a motion dataset, such as the INTERACTION \cite{zhan_interaction_2019}, TAF \cite{zipfl_traffic_2020}, or inD \cite{bock_ind_2019} dataset, and makes it available in a suitable form so that criticality values can then be calculated using the corresponding metrics.

The data basis is traffic scenarios discretized by scenes. Each scene, i.e., snapshot at a given time, is defined by the states of the traffic participants (entities) in it at that time. That is, each traffic participant at a discrete point in time is represented by its pose, velocity, acceleration, size, and classification.

The actual metrics computation takes place in parallel and independently of each other. Since the calculation of some metrics takes an exceptionally long time but also partly requires "a posterior" knowledge of the scenario, the framework in its full depth can only be carried out offline, i.e., after recording the entire scenario in case of microscopic metrics.

Depending on the metric type, several values may be calculated for a scene, for example, for each traffic participant separately (compare nanoscopic metrics). These are mapped to one value by a permutation-invariant function (e.g., $max, mean, min$).

Most distance- and time-based metrics go towards zero when a scene is critical. For example, a $\mathrm{TTC} = 0.0$ indicates a collision took place.
Furthermore, most metrics have a decreasing impact with an increasing value.

To compare all the previously mentioned metrics, including traffic quality and safety potential with each other, we use the following function to invert and scale all used metrics in case they do not already return values between $0$ (non-critical) and $1$ (critical):
\begin{equation}
    f(x) = e^{\alpha(-x)}
\end{equation}
where $x$ is the measured metric value and $\alpha=1.0$ a coefficient regarding the sensitivity.
The coefficient $\alpha=1.0$ in the exponent can be replaced with a more suitable coefficient in case a metric needs to be less sensitive (e.g. $\alpha=2.0$) or more sensitive (e.g. $\alpha=0.5$).
Finally, all calculated metrics are compared to a uniform visualization and consolidated into one graph.


\section{Metric Implementation}
In this section, the implementation of the new improved metrics (traffic quality and safety potential) are described.
\label{sec:implementation}
\subsection{Inverse universal Traffic Quality}
The inverse universal traffic quality (TQ) is based on the traffic quality for highway scenarios as proposed by Hallerbach \textit{et al.} \cite{hallerbach_simulation-based_2018}.
 In contrast, no final score is calculated where each submetrics is weighted by a map-dependent score for the inverse universal traffic quality.
 Moreover, we use the term inverse since measured values rise with the growing criticality of a scene and, therefore, its quality is decreasing.
 The four sub-metrics are used to span an area that grows larger with decreasing quality and criticality.
 Additionally, information from previous time frames is avoided as much as possible to simplify computation.

 At the center of the metric is a road user, the ego vehicle, to which all measured and calculated values refer.
 The three scenes in \Cref{fig:maps} show different maps and traffic situations.

 \subsubsection{Macroscopic Inverse Traffic Quality}
 The macroscopic traffic quality computes the coefficient of variation with respect to the velocities of all vehicles in a scene, i.e., the higher the macroscopic value, the more unsteadily all road users move.
 A steady traffic is marked as uncritical,
 Uniform traffic is considered uncritical, regardless of the average speed of all participants.
 The coefficient for a complete scene is calculated with
 \begin{equation}
     \mathrm{TQ}_{macro} = \frac{\sigma_{scene}}{\bar{\nu}_{scene}}
 \end{equation}
 where $\sigma_{scene}$ describes the standard deviation of the velocity and $\bar{\nu}_{scene}$ the mean velocity of all vehicles within one scene.
 The macroscopic value is the same for all traffic participants within one scene.
 \subsubsection{Microscopic Universal Traffic Quality}
 The microscopic traffic quality is a coefficient between all vehicles on the map and the ones within the estimated braking distance of an ego car $A$:
 
  \begin{equation}
     \mathrm{TQ}_{micro} = \frac{\sum_{A \in Veh_t}^{}\chi_{brake}(A)}{\sum_{A \in Veh_t}\chi_{scene}(A)}
 \end{equation}
 where $\chi_{brake}$ and $\chi_{scene}$ denote the vehicle count within the braking distance and the complete scene, respectively, and $Veh_t$ the set of all vehicles found on the map at time $t$.

 \subsubsection{Nanoscopic Inverse Traffic Quality}
 The nanoscopic traffic quality is similar to the macroscopic traffic quality. However, it only considers the coefficient of variation within the braking distance of an ego vehicle
  \begin{equation}
     \mathrm{TQ}_{nano} = \frac{\sigma_{brake}}{\bar{\nu}_{brake}}
 \end{equation}
  where $\sigma_{brake}$ describes the standard deviation of the speed and $\bar{\nu}_{brake}$ the mean velocity of all vehicles within the braking distance of the ego vehicle.
 
 \subsubsection{Individual Inverse Traffic Quality}
 The individual traffic quality is the only submetric that considers data from previous time frames, and it only considers the ego velocity and acceleration development over the last few seconds.
 It consists of two coefficients combined
   \begin{equation}
     \mathrm{TQ}_{indi} = \frac{\frac{\bar{a}_{ego}}{a_{ref}} + \frac{\bar{\nu}_{ego}}{\nu_{ref}}}{2}
 \end{equation}
 where $\bar{a}_{ego}$ describes the mean acceleration and $\bar{\nu}_{ego}$ the mean velocity of the ego vehicle over a certain time in the past.
 Values $a_{ref}$ and $\nu_{ref}$ are reference values, e.g., $\nu_{ref} = 50 \frac{km}{h}$ for urban traffic.
 
\subsubsection{Traffic Quality in Different Scenes}

\begin{table}[tbp]
\centering
\captionsetup{justification=centering}
\caption{All maximum traffic quality values for scenes in \Cref{fig:maps}}
\begin{tabular}{@{}lllll@{}}
\toprule
                & Macro     & Micro     & Nano      & Indi      \\ \midrule
Roundabout      & 0.5634    & 0.375     & 0.7097    & 0.5702    \\
Merging         & 0.7573    & 0.8286    & 0.7532    & 0.6081    \\
Intersection    & 1.0254    & 0.4118    & 1.4091    & 0.3996    \\ \bottomrule
\label{tab:traffic_quali}
\end{tabular}
\vspace{-3ex}
\end{table}

\Cref{tab:traffic_quali} shows a list with results for all four submetrics, measured from the scenes in \Cref{fig:maps}.

In general, the four submetrics bring the following general findings in a scene:
\begin{itemize}
    \item Macroscopic traffic quality scores higher if the vehicle movement on the whole map is unsteady, e.g., many vehicles have to brake or are accelerating. A constant flow or only stationary traffic objects have a low macroscopic score close to $0$.
    \item Microscopic traffic quality scores higher if a scene shows high traffic density and fast-moving vehicles. In case vehicles have more space to react, the microscopic score decreases.
    \item Nanoscopic traffic quality scores higher if the vehicle movement of all traffic participants close to the ego vehicle is unsteady, e.g., one or more vehicles are reacting to something in its near vicinity.
    \item Individual traffic quality scores higher if the ego vehicle moves faster than a given reference value or has a higher acceleration than the typical reference values, e.g., an ego vehicle moving faster than a reference value of $50 \frac{km}{h}$ or performing an emergency brake.
\end{itemize}

\subsection{Safety Potential}
The safety potential (SP) $\rho$ describes the state of a traffic scene with regard to possible collisions of traffic participants, based on the current state of all traffic participants. 
This is an intermediate result in the SFF model and is used to calculate an appropriate action for the ego vehicle. In this work, we will exploit the assessing characteristic of this safety potential to examine the criticality of relationships between traffic participants and ultimately create an assessment metric for a traffic scene.

In order to reduce the computation effort of the continuous safety potential, we only consider possible overlaps of claimed sets of two traffic participants. Suppose there are no overlaps in a scene, $\rho$ equals zero. Furthermore, the separate consideration allows the individual evaluation of all traffic participants in the scene.
\begin{figure}[htbp]
  \centering
     \includegraphics[width=0.90\linewidth]{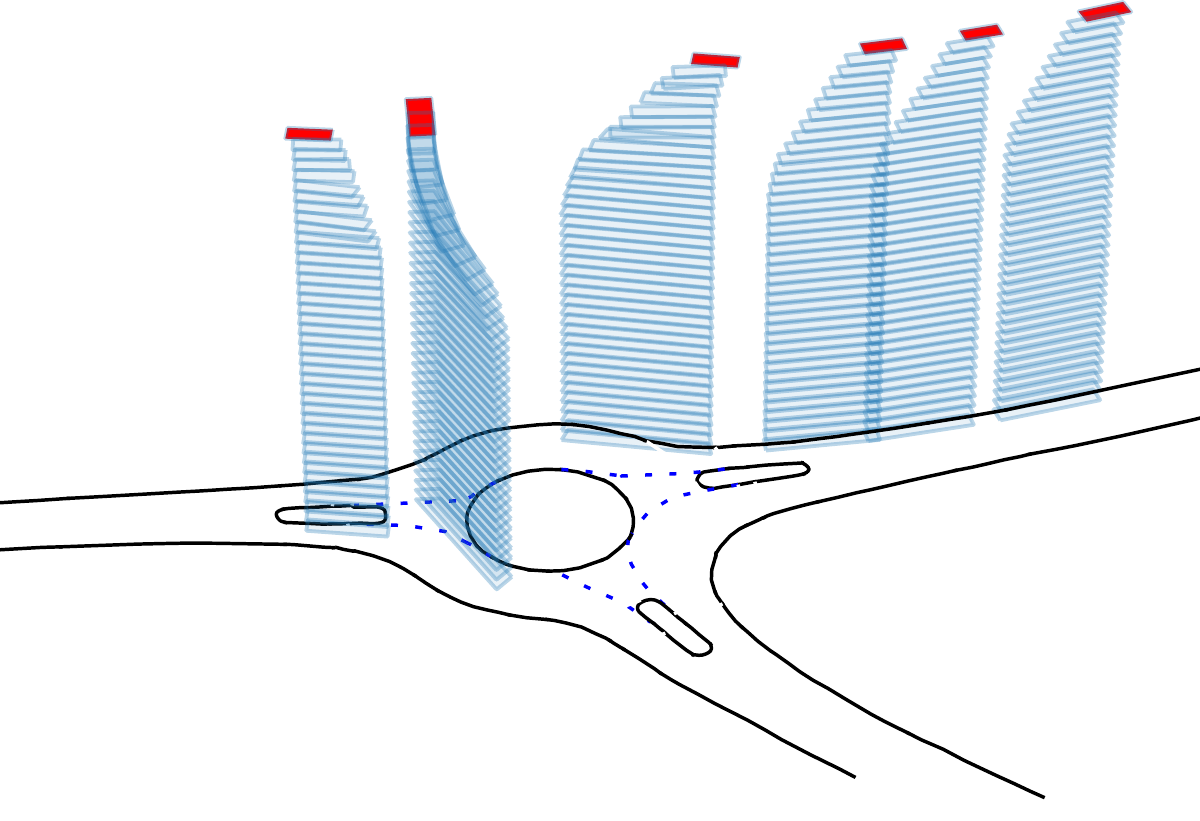}
  \caption{Claimed sets (blue rectangles) of all traffic participants in the scene.}
  \label{fig:claimed-set}
\end{figure}

The calculation of the safety potential consists of three steps. Starting with calculating the safety procedure $S_A$ of all traffic participants in the scene. 
The state $X_A$ of a traffic participant $A$ is described by its position $s_A$ and velocity $\nu_A$ to a given time $t$, discretized by a given step width in the time horizon $T$.
In our case, we only consider a simple safety procedure, where the actor slows down by braking with acceleration $a_{min}$. This results in
\begin{align}
    s_A(t) = \nu_A(0) \cdot t + a_{min} \frac{t^2}{2}
\end{align}
\begin{align}
    \nu_A(t) = max\big(\nu_A(0) \cdot t + a_{min} \cdot t , 0 \big).
\end{align}

Since, in addition to the fastest possible deceleration, we also want to represent slower braking maneuvers, we define $S'_A$ for the states that would occur in a safety procedure with reaction time $t_r$ and lower braking deceleration $a'$. The position states $s'_A$, and the velocity states $\nu'_A(t)$ are then defined as follows:

\begin{align}
    s'_A(t) = \nu_A(0) \cdot t + a' \cdot \frac{max(t-t_r,0)^2}{2} \notag\\+ \ min(t,t_r) \cdot \nu_A(0)
\end{align}
\begin{align}
    \nu'_A(t) = \nu_A(0) + a' \cdot max(t-t_r)
\end{align}

To generate a realistic braking behavior, the positions $s_A(t) \in S_A$ are not calculated in a straight line based on the last given pose of $A$, but along the course of the road (see second vehicle from the left in \Cref{fig:claimed-set}). 

\begin{figure}[htbp]
    \centering
    \includegraphics[width=0.80\linewidth]{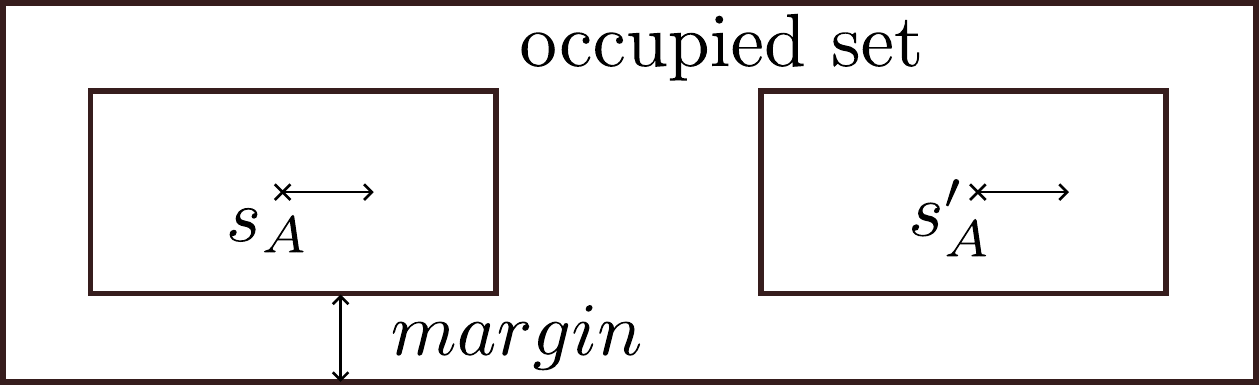}
    \caption{Occupied set of a traffic participant $A$}
    \label{fig:occupied-set}
\end{figure}

The next step is calculating the claimed set, namely the space in the time horizon, which is captured by the vehicle's future trajectory.
The occupied set, i.e., the space occupied by a vehicle at a given time $t$, is described by a polygon (see \Cref{fig:occupied-set}).

The vertices of the polygon are spanned by the two anchor points of the positions of $s_A$ and $s'_A$ and the respective length and width of the vehicle plus a certain safety margin.
The occupied set is therefore defined by the area between the best possible braking maneuver $S_A$ and a realistic braking maneuver $S'_A$.
In \Cref{fig:claimed-set} the occupied sets of all six traffic participants are shown graphically. 
The red rectangle represents the state of the current timestamp $t=0$. The temporal development (here $T=4\,sec$) is shown on the z-axis ($0<t \leq T$) (later states are plotted lower).

An essential part of calculating the safety potential is the time required for a vehicle to decelerate to a standstill $t^{stop}_A$ in each state.
\begin{align}
    t^{stop}_A(t)= \frac{\nu'_t}{a'} - max(t_r-t,0)
\end{align}
With the help of $t^{stop}_A$, the future state (in the claimed set) can be additionally rated. For example, if one vehicle collides with another at a low speed, this is not considered as severe as if the two vehicles collide at a very high speed.

Finally, the overlapping area $\mathcal{A}^{AB}(t)$ of the occupied sets of all traffic participants is calculated at each time $t$. Afterwards, $\mathcal{A}^{AB}(t)$ is then weighted by $t^{stop}_A(t)$.

This results in an actor specific safety potential $\rho_{AB}$ between actor $A$ and actor $B$.
\begin{align}
    \rho_{AB} = \sum_t \mathcal{A}^{AB}(t) \cdot t_A^{stop}(t)
\end{align}

To make the values of $\rho$ comparable to other criticality metrics, they are mapped to an interval between $0$ and $1$ by a scaled $tanh$ function.

%% file: content/05_evaluation.tex
\section{Experiments and Evaluation}
\label{sec:evaluation}
\begin{figure*}[htbp]
    \centering
       \includegraphics[width=\linewidth]{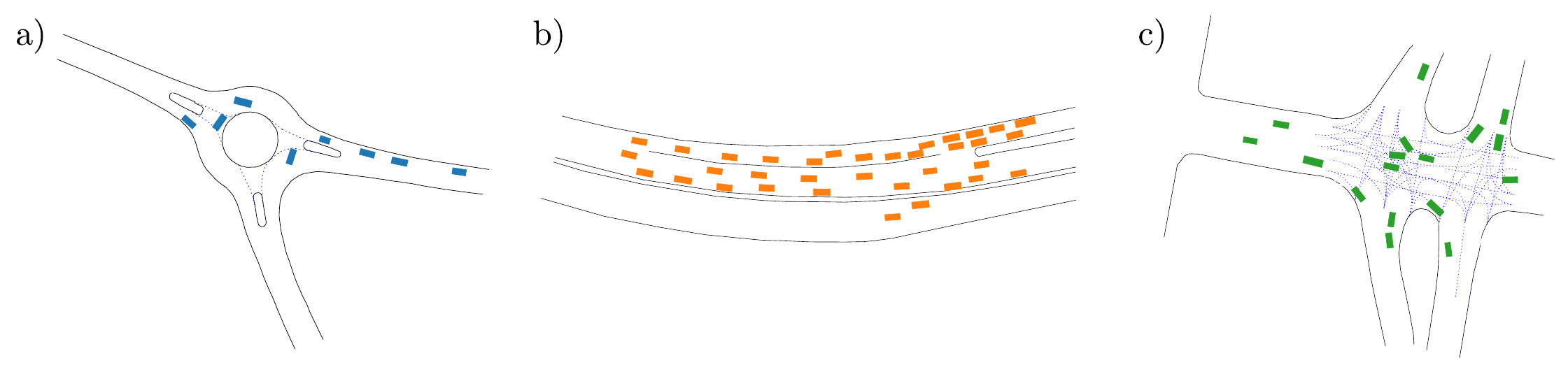}
\caption{Scenes from different Interaction dataset scenarios: roundabout (blue traffic participants, (a)), merging (orange traffic participants, (b)), intersection, green traffic participants, (c))}
\label{fig:maps}
\end{figure*}
For an exemplary evaluation, we use three different scene types. These include a roundabout scene (see \Cref{fig:maps}a), a merging scene on a highway (\Cref{fig:maps}b), and an urban intersection (\Cref{fig:maps}c).

Except for the three cars passing the roundabout, most vehicles are standing still and waiting for an opportunity to enter the roundabout or decelerate to avoid a collision.
The merging scene (\Cref{fig:maps}b) has a high traffic density with little space around some actors and can be seen as a critical situation.
The intersection scene (\Cref{fig:maps}c), indicated as one of the high risk situations by Zhan \textit{et al.}  \cite{zhan_interaction_2019}.
Three cars in opposite directions are crossed by one car that passes in between them.
Most other vehicles have low values since most are standing and waiting to enter the intersection or are approaching it.

\subsection{Inv. Universal Traffic Quality and Safety Potential}
The enhancements of safety potential and traffic quality proposed in this paper are intended to address the limitation that metrics can often only be applied to specific types of scenes. The traffic quality focuses more on the speed variance in different zoom levels and the safety potential of possible collisions in the geometric space. 

If we look at the safety potential in different traffic scenes, we find that it has a value greater than 0 for critical scenes, regardless of the type. Of course, if the scene is irrelevant in the context of criticality (see the violet scenario in \Cref{fig:spider3}), the safety potential also becomes zero.
The same characteristics also apply to the traffic quality metric. In general, all traffic scene types investigated can also be represented here. Furthermore, these are described by four values instead of a single value and give more insight into the current depicted scene and ego vehicle.
The macroscopic value evaluates the overall scene regarding its vehicle movement continuity, whereas microscopic traffic quality compares the surrounding of the ego vehicle to the complete scene with respect to the ego vehicle's velocity. A larger part of the scene is considered if the ego vehicle moves faster while standing still or slow movement leads to smaller areas of consideration. The nanoscopic traffic quality describes the criticality in the ego vehicle's close neighborhood, and the individual score only considers the ego vehicle's behavior.

\subsection{Multidimensional Metric Assessment}
\label{sec:multi_metric}
\begin{figure*}[htbp]
\newcommand{\scalemykiviat}{0.73} 
\centering

\begin{subfigure}[t]{.3\textwidth}
        \centering
        \raisebox{8.2mm}{
        \scalebox{\scalemykiviat}{
        \trimbox{1.2cm 0mm 0.9cm 0mm}{
        \input{figures/kiviat_spider_complete.tikz}}}}
        \caption{Comparison of the scenes in \Cref{fig:maps}}
        \label{fig:spider1}
    \end{subfigure}%
    \quad
    \begin{subfigure}[t]{0.3\textwidth}
        \centering
        \scalebox{\scalemykiviat}{
        \trimbox{1.2cm 0mm 0.9cm 0mm}{
        \input{figures/kiviat_scene_types.tikz}}}
        \caption{Comparison between the domain-dependent metrics}
        \label{fig:spider2}
    \end{subfigure}
    \quad
    \begin{subfigure}[t]{0.3\textwidth}
        \centering
        \raisebox{8.2mm}{
        \scalebox{\scalemykiviat}{
        \trimbox{1.2cm 0mm 0.9cm 0mm}{
        \input{figures/kiviat_criticalities.tikz}}}}
        \caption{Comparison of a critical (green) and an uncritical (violet) scene}
        \label{fig:spider3}
    \end{subfigure}
    \vspace{1ex}
    \caption{Macro, Micro, Nano, Indi: Traffic Quality, TD: Trajectory Distance, GT: Gap Time, ET: Encroachment Time. PET Post-Encroachment Time, SP: Safety Potential, WTTC: Worst-Time-To-Collision, Dist: Euclidean Distance, TTC: Time-To-Collision.}
    \label{fig:all_spiders}
\end{figure*}
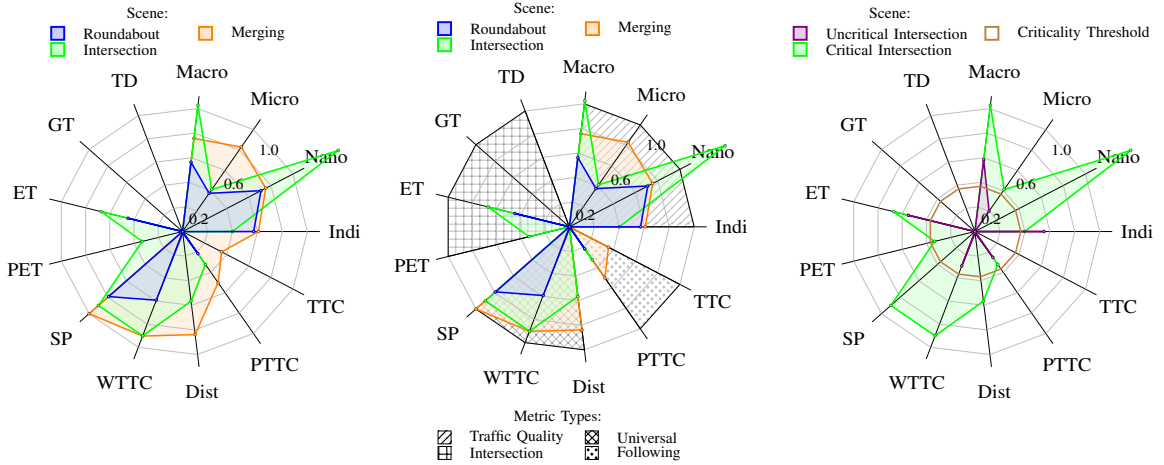

The metric evaluation of the proposed illustration with Kiviat diagrams (\Cref{fig:all_spiders}) offers different possibilities. We call the composite of different criticality measures into one graphical representation the fingerprint of a scene.

\Cref{fig:spider1} shows a diagram of all three scenes' overall proposed metrics.
The larger the spanned area between the individual metrics, the more critical the scene.

The area of the merging scenario is missing data points for all metrics indicating an intersection. Therefore, it can be assumed that no relevant situation with intersecting trajectories between any two vehicles exists in this scene.
Since it is the orange merging scenario of \Cref{fig:maps}b), it is possible to see that this is indeed the case.
Additionally, the distribution of the fingerprint can be used to identify the scene type (compare \Cref{tab:criticality_area}). This could open up new possibilities for automated clustering of scenarios and simultaneously assigning a classification.

Not all scenes have data values larger than zero for every axis. This can have two reasons: The metric is not critical in this scene, or this specific metric does not support the investigated scene type. The latter phenomenon is investigated in more detail in \Cref{fig:spider2}. Metrics stemming from the same domains have the requirements or the same use cases are combined, and their joint area is marked graphically by a pattern (Traffic Quality, Intersection, Universal, Following).
This establishes features in the Kiviat diagram analogous to features found in fingerprints, e.g., loops, arches, and windings.
Not every scenario possesses every feature, and each scene has different manifestations of the found features.

The roundabout scene (blue) from \Cref{fig:maps}a) only has values for PET and ET in both Kiviat diagrams in \Cref{fig:spider1} and \Cref{fig:spider2}. 
This indicates that there are intersections between some of the currently moving vehicles.
However, since both are calculated for a scenario and not for a scene, the constellation of cars shown in \Cref{fig:maps}a) takes place after the passing of the point or area of intersection.
Generally, the metric groups' areas are comparatively small, indicating a non-critical scene.
As previously mentioned, the merging scene (orange) shows no intersection metric, which leads to the conclusion that there is no intersection found.
However, it has the highest score for the car-following metrics for all three scenes.
The last scene is a classical intersection (green) and is shown in \Cref{fig:maps}c).
This scene includes all proposed metric groups and is evaluated as more critical than the roundabout scene when considering the investigated metrics.

\Cref{fig:spider3} shows metrics for the scene in \Cref{fig:maps}c) (green) and the same map at a different time step with only five cars (violet) from the same dataset and a possible threshold (brown). 
Since five cars can drive independently, this scene shows a low criticality for all metrics.
The brown line indicates a criticality threshold of $1.5sec$ or $1.5m$ as proposed by Allen \textit{et al.} \cite{allen_analysis_1978}. 
It is used for non-temporal- or distance-based metrics as well to get a complete circle but can be replaced with thresholds regarding the use case or requirements for each metric.
The area of this threshold circle can then be compared with the area of both scenes to define if a scene is critical.

In order to make scenes easier to compare despite their multidimensionality, we make use of the previous statement that more critical scenes have a larger area between the individual criticality metrics in the Kiviat diagram.
This makes the geometric area a holistic criticality measure and allows scenes to be broadly classified. 
Values for the demonstrated, exemplary scenes are shown in \Cref{tab:criticality_area}.
Each column shows the criticalities for each scenario (roundabout, merging, intersection critical, intersection non-critical), where the first row is the value of the overall criticality as depicted in \Cref{fig:spider1} and the following rows are the domain-dependent values.

\begin{table}[tbp]
\centering
\captionsetup{justification=centering}
\caption{Areas of criticality for \Cref{fig:spider1} in the first row and the domain-dependent values of \Cref{fig:spider2} in the following rows}
\begin{tabular}{@{}llllll@{}}
\toprule
             & Round.& Merg.&  Inters.c &  Inters.n \\ \midrule
Compl.        & 0.3074  & 0.9713  & 0.7018 & 0.0806 \\
TQ            & 0.2049& 0.3972& 0.3637 & 0.027 \\
Intersection  & 0.0   & 0.0   & 0.0518 & 0.0 \\
Universal     & 0.1107& 0.3853& 0.3069 & 0.0008 \\
Following      & 0.0001&0.0412 &0.0001 &0.0\\
\bottomrule
\end{tabular}
\label{tab:criticality_area}

\end{table}

\subsection{Metric Performance}
In order to evaluate the metrics we proposed, namely Traffic Quality and Safety Potential, these metrics are compared to conventional metrics. For this purpose, we use TTC as a representative metric for all following scenes and PET as a representative metric for all scenes with intersecting situations as ground truth.
\Cref{tab:glob_eval} shows the results of the metrics at three different road topologies. 
If one of the ground truth metrics (TTC, PET) exceeds the threshold, the scene is considered critical. With this assumption, the sensitivity and specificity (true negative TN, true positive TP, false positive FP, and false negative FN) can be determined for SP and TQ for the examined dataset.
Both presented metrics show similar properties regarding the evaluation of different road topologies. For both metrics, the roundabout scenario has very high specificity but very low sensitivity. In the intersection and merging scenario, TQ has high sensitivities.
In summary, it can be said that our metrics approximate the criticality description of TTC and PET particularly well, especially for intersection scenarios. 
In general, however, it must be emphasized that TTC and PET do not necessarily serve as ground truth for evaluating all traffic scenes (see above).

\begin{table}[tbp]
\centering
\captionsetup{justification=centering}
\caption{Evaluation of the Safety Potential and the universal inverse Traffic Quality metric in comparison with TTC and PET}
\begin{tabular}{@{}lllll@{}}
\toprule
                    &       & Roundabout      & Intersection      & Merging             \\ \midrule
\multirow{6}{*}{Safety Potential} & TP  & 0.0       & 0.11    & 0.86      \\
                    & TN    & 0.89     & 0.48     & 0.0        \\
                    & FP    & 0.09    & 0.25     & 0.14     \\
                    & FN    & 0.02      & 0.16     & 0.0         \\
                    & Sens  & 0.0       & 0.41      & 1.0      \\
                    & Spec  & 0.91      & 0.65      & 0.0      \\\midrule
\multirow{6}{*}{Traffic Quality} & TP  & 0.001   & 0.23   & 0.75      \\
                    & TN    & 0.95     & 0.21    & 0.05     \\
                    & FP    & 0.02      & 0.52   & 0.11     \\
                    & FN    & 0.03      & 0.04    & 0.1      \\
                    & Sens  & 0.05      & 0.86      & 0.87      \\
                    & Spec  & 0.97      & 0.29      & 0.29      \\\midrule
crit.               &       & 0.02    & 0.27   & 0.14          \\
non-crit.           &       & 0.98   & 0.73   & 0.86         \\ \bottomrule
\end{tabular}
\label{tab:glob_eval}
\end{table}

%% file: figures/kiviat_spider_complete.tikz
\begin{tikzpicture}[
  bluenode/.style={shape=rectangle, draw=blue, fill=blue!20, line width=1},
  greennode/.style={shape=rectangle, draw=green, fill=green!20, line width=1},
  orangenode/.style={shape=rectangle, draw=orange, fill=orange!20, line width=1},
groupnode/.style = {shape=rectangle, fill=#1!50, line width=0.4},
block/.style={signal, font=\footnotesize, signal pointer angle=140, align=center}]

\tkzKiviatDiagram[scale=.45,label distance=0.2cm,
        gap     = 1,  
        lattice = 5,
        radial style/.style ={-}
        ]{Indi, Nano, Micro, Macro, TD, GT, ET, PET, SP, WTTC, Dist, PTTC, TTC}
\tkzKiviatLine[thick,
                mark=ball,
                ball color=orange,
                color=orange,
                fill=orange!20,
                opacity=.5]%
                (5*0.6081,5*0.7532,5*0.8286,5*0.7573,0.0,0.0,0.0,0.0,5*1.0,5*0.9022,5*0.8384,5*0.5016,5*0.3535) 
\tkzKiviatLine[thick,
                mark=ball,
                fill=green!20,
                ball color=green,
                color=green]%
                (5*0.3996,5*1.4091,5*0.4118,5*1.0254,5*0.0009,0.0,5*0.6703,5*0.3329,5*0.8999,5*0.901,5*0.5661,5*0.323,5*0.0019) 
\tkzKiviatLine[thick,
                color=blue,
                mark=ball,
                ball color=blue,
                fill=blue!20,
                opacity=.5]%
                (5*0.5702,5*0.7097,5*0.375,5*0.5634,0.0,0.0,5*0.4493,5*0.0002,5*0.7918,5*0.5929,5*0.0119,5*0.2165,5*0.0021)

\node[block, minimum height=65pt, minimum width=60pt, signal from=west, anchor=west] at (-0.8, 0.5) (t02) {\footnotesize 0.2};
\node[block, minimum height=65pt, minimum width=60pt, signal from=west, anchor=west] at ($(t02.west) + (up:40pt) + (right:40pt)$) (t06) {\footnotesize 0.6};
\node[block, minimum height=65pt, minimum width=60pt, signal from=west, anchor=west] at ($(t06.west) + (up:40pt) + (right:40pt)$) (t10) {\footnotesize 1.0};

  \node [bluenode,label=right:\footnotesize Roundabout] at (-5., 8.0) (l1) {}; 
  \node [greennode,label=right:\footnotesize Intersection] at ($(l1.south) + (down:10pt) + (right:0pt)$)  (l2) {}; 
  \node [orangenode,label=right:\footnotesize Merging] at ($(l1.east) + (down:0pt) + (right:160pt)$)  (l3){}; 
  \node [groupnode=white,label=right:\footnotesize Scene:] at ($(l1.north) + (up:15pt) + (right:50pt)$) (l4) {}; 
\end{tikzpicture}

%% file: figures/kiviat_scene_types.tikz
\begin{tikzpicture}[
  bluenode/.style={shape=rectangle, draw=blue, fill=blue!20, line width=1},
  greennode/.style={shape=rectangle, draw=green, fill=green!20, line width=1},
  orangenode/.style={shape=rectangle, draw=orange, fill=orange!20, line width=1},
    groupnode/.style = {shape=rectangle, fill=#1!50, line width=0.4},
block/.style={signal, font=\footnotesize, signal pointer angle=140, align=center}]

\tkzKiviatDiagram[
        scale=.45,label distance=0.2cm,
        gap     = 1,  
        lattice = 5,
        radial style/.style ={-}
        ]{Indi, Nano, Micro, Macro, TD, GT, ET, PET, SP, WTTC, Dist, PTTC, TTC}
\tkzKiviatLinePattern[
                fill=black,
                opacity=.5,
                pattern=north east lines
                ]%
                (5,5,5,5,0.0,0.0,0.0,0.0,0.0,0.0,0.0,0.0,0.0)
\tkzKiviatLinePattern[
                fill=black,
                opacity=.5,
                pattern=grid
                ]%
                (0.0,0.0,0.0,0.0,5,5,5,5,0.0,0.0,0.0,0.0,0.0) 
\tkzKiviatLinePattern[
                fill=black,
                opacity=.5,
                pattern=crosshatch
                ]%
                (0.0,0.0,0.0,0.0,0.0,0.0,0.0,0.0,5,5,5,0.0,0.0)  
\tkzKiviatLinePattern[
                fill=black,
                opacity=.5,
                pattern=crosshatch dots
                ]%
                (0.0,0.0,0.0,0.0,0.0,0.0,0.0,0.0,0.0,0.0,0.0,5,5)  

\tkzKiviatLine[thick,
                mark=ball,
                ball color=orange,
                color=orange,
                fill=orange!20,
                opacity=.5]%
                (5*0.6081,5*0.7532,5*0.8286,5*0.7573,0.0,0.0,0.0,0.0,0.0,0.0,0.0,0.0,0.0)
\tkzKiviatLine[thick,
                mark=ball,
                ball color=orange,
                color=orange,
                fill=orange!20,
                opacity=.5]%
                (0.0,0.0,0.0,0.0,0.0,0.0,0.0,0.0,0.0,0.0,0.0,0.0,0.0) 
\tkzKiviatLine[thick,
                mark=ball,
                ball color=orange,
                color=orange,
                fill=orange!20,
                opacity=.5]%
                (0.0,0.0,0.0,0.0,0.0,0.0,0.0,0.0,5*1.0,5*0.9022,5*0.8384,0.0,0.0)  
\tkzKiviatLine[thick,
                mark=ball,
                ball color=orange,
                color=orange,
                fill=orange!20,
                opacity=.5]%
                (0.0,0.0,0.0,0.0,0.0,0.0,0.0,0.0,0.0,0.0,0.0,5*0.5016,5*0.3535)                 
\tkzKiviatLine[thick,
                mark=ball,
                fill=green!20,
                ball color=green,
                color=green]%
                (5*0.3996,5*1.4091,5*0.4118,5*1.0254,0.0,0.0,0.0,0.0,0.0,0.0,0.0,0.0,0.0)
\tkzKiviatLine[thick,
                mark=ball,
                fill=green!20,
                ball color=green,
                color=green]%
                (0.0,0.0,0.0,0.0,5*0.0009,0.0,5*0.6703,5*0.3329,0.0,0.0,0.0,0.0,0.0) 
\tkzKiviatLine[thick,
                mark=ball,
                fill=green!20,
                ball color=green,
                color=green]%
                (0.0,0.0,0.0,0.0,0.0,0.0,0.0,0.0,5*0.8999,5*0.901,5*0.5661,0.0,0.0)  
\tkzKiviatLine[thick,
                mark=ball,
                fill=green!20,
                ball color=green,
                color=green]%
                (0.0,0.0,0.0,0.0,0.0,0.0,0.0,0.0,0.0,0.0,0.0,5*0.323,5*0.0019)  
\tkzKiviatLine[thick,
                color=blue,
                mark=ball,
                ball color=blue,
                fill=blue!20,
                opacity=.5]%
                (5*0.5702,5*0.7097,5*0.375,5*0.5634,0.0,0.0,0.0,0.0,0.0,0.0,0.0,0.0,0.0)
\tkzKiviatLine[thick,
                color=blue,
                mark=ball,
                ball color=blue,
                fill=blue!20,
                opacity=.5]%
                (0.0,0.0,0.0,0.0,0.0,0.0,5*0.4493,5*0.0002,0.0,0.0,0.0,0.0,0.0) 
\tkzKiviatLine[thick,
                color=blue,
                mark=ball,
                ball color=blue,
                fill=blue!20,
                opacity=.5]%
                (0.0,0.0,0.0,0.0,0.0,0.0,0.0,0.0,5*0.7918,5*0.5929,5*0.0119,0.0,0.0)  
\tkzKiviatLine[thick,
                color=blue,
                mark=ball,
                ball color=blue,
                fill=blue!20,
                opacity=.5]%
                (0.0,0.0,0.0,0.0,0.0,0.0,0.0,0.0,0.0,0.0,0.0,5*0.2165,5*0.0021)  
\node[block, minimum height=65pt, minimum width=60pt, signal from=west, anchor=west] at (-0.8, 0.5) (t02) {\footnotesize 0.2};
\node[block, minimum height=65pt, minimum width=60pt, signal from=west, anchor=west] at ($(t02.west) + (up:40pt) + (right:40pt)$) (t06) {\footnotesize 0.6};
\node[block, minimum height=65pt, minimum width=60pt, signal from=west, anchor=west] at ($(t06.west) + (up:40pt) + (right:40pt)$) (t10) {\footnotesize 1.0};

  \node [bluenode,label=right:\footnotesize Roundabout] at (-5., 8.0) (l1) {}; 
  \node [greennode,label=right:\footnotesize Intersection] at ($(l1.south) + (down:10pt) + (right:0pt)$)  (l2) {}; 
  \node [orangenode,label=right:\footnotesize Merging] at ($(l1.east) + (down:0pt) + (right:160pt)$)  (l3){}; 
  \node [groupnode=white,label=right:\footnotesize Scene:] at ($(l1.north) + (up:15pt) + (right:50pt)$) (l4) {}; 
  
    \node [groupnode=black,label=right:\footnotesize Traffic Quality, pattern=north east lines,draw=black] at (-5., -8.5) (l7) {}; 
    \node [groupnode=white,label=right:\footnotesize Metric Types:] at ($(l7.north) + (up:15pt) + (right:50pt)$) (l6) {}; 
  \node [groupnode=black,label=right:\footnotesize Intersection, pattern=grid,draw=black] at ($(l7.south) + (down:10pt) + (right:0pt)$)  (l8) {}; 
  \node [groupnode=black,label=right:\footnotesize Universal, pattern=crosshatch,draw=black] at ($(l7.east) + (down:0pt) + (right:160pt)$)  (l9){}; 
  \node [groupnode=black,label=right:\footnotesize Following, pattern=crosshatch dots,draw=black] at ($(l8.east) + (down:0pt) + (right:160pt)$)  (l10){}; 

\end{tikzpicture}

%% file: figures/kiviat_criticalities.tikz
\begin{tikzpicture}[
  tealnode/.style={shape=rectangle, draw=violet, fill=violet!20, line width=1},
  brownnode/.style={shape=rectangle, draw=brown, line width=1},
  greennode/.style={shape=rectangle, draw=green, fill=green!20, line width=1},
  groupnode/.style = {shape=rectangle, fill=#1!50, line width=0.4},
block/.style={signal, font=\footnotesize, signal pointer angle=140, align=center}]

\tkzKiviatDiagram[scale=.45,label distance=0.2cm,
        gap     = 1,  
        lattice = 5,
        radial style/.style ={-}
        ]{Indi, Nano, Micro, Macro, TD, GT, ET, PET, SP, WTTC, Dist, PTTC, TTC}
\tkzKiviatLine[thick,
                mark=ball,
                fill=green!20,
                ball color=green,
                color=green]%
                (5*0.3996,5*1.4091,5*0.4118,5*1.0254,5*0.0009,0.0,5*0.6703,5*0.3329,5*0.8999,5*0.901,5*0.5661,5*0.323,5*0.0019) 
\tkzKiviatLine[thick,
                mark=ball,
                ball color=violet,
                color=violet,
                fill=violet!20,
                opacity=.5]%
                (5*0.5543,0.0,5*0.2,5*0.5824,5*0.0,5*0.0004,5*0.5488,5*0.0,0.0,5*0.297,5*0.0118,5*0.2541,0.0) 
                
\tkzKiviatLine[thick,
                color=brown,
                opacity=.5]%
                (5*0.3678,5*0.3678,5*0.3678,5*0.3678,5*0.3678,5*0.3678,5*0.3678,5*0.3678,5*0.3678,5*0.3678,5*0.3678,5*0.3678,5*0.3678)

\node[block, minimum height=65pt, minimum width=60pt, signal from=west, anchor=west] at (-0.8, 0.5) (t02) {\footnotesize 0.2};
\node[block, minimum height=65pt, minimum width=60pt, signal from=west, anchor=west] at ($(t02.west) + (up:40pt) + (right:40pt)$) (t06) {\footnotesize 0.6};
\node[block, minimum height=65pt, minimum width=60pt, signal from=west, anchor=west] at ($(t06.west) + (up:40pt) + (right:40pt)$) (t10) {\footnotesize 1.0};

  \node [tealnode,label=right:\footnotesize Uncritical Intersection] at (-7., 8.) (l1) {}; 
  \node [greennode,label=right:\footnotesize Critical Intersection ] at ($(l1.south) + (down:10pt) + (right:0pt)$)  (l2) {}; 

  \node [groupnode=white,label=right:\footnotesize Scene:] at ($(l1.north) + (up:15pt) + (right:50pt)$) (l4) {}; 
    \node [brownnode,label=right:\footnotesize Criticality Threshold] at ($(l1.east) + (down:0pt) + (right:210pt)$)  (l3){}; 
\end{tikzpicture}

%% file: content/06_conclusion.tex
\section{Conclusion and Outlook}
\label{sec:conclusion}
In this work, we proposed two improved metrics suitable for evaluating complete traffic scenes in scenario-based testing of automated and highly automated driving systems.
The inverse universal traffic quality is based on a previous approach to traffic quality \cite{hallerbach_simulation-based_2018}, and the safety potential is based on an 
interim result of the safety force field \cite{nister_safety_2019}.

Further, these metrics were used for generalized evaluation purposes combined with the established criticality metrics for different scenario types, e.g., intersection or car-following scenario metrics, and completed with different scenario type independent metrics.

We used Kiviat diagrams for evaluation and showed that the spanned area could be used as a score for an overall criticality. 
The spanned area can be computed for all metrics or a subset of metrics, depending on the use case for which the evaluation is needed.

In the future, the examined metrics will be adapted to different road geometries. In all metrics used, the road courses are not taken into account apart from the safety potential.
Furthermore, in our experiments, no distinction was made between vulnerable road users and vehicles. However, especially in urban intersections, scenes with pedestrians are particularly critical and should be considered separately.
In addition, the descriptive nature of the scene fingerprint could be used to automate the clustering and classification of large sets of traffic scenarios for scenario-based testing.